\documentclass[runningheads]{llncs}
\usepackage{graphicx}
\usepackage{amsmath,amssymb} 
\usepackage{color}
\usepackage{hyperref}
\usepackage{subcaption}
\usepackage[labelfont=bf]{caption}
\graphicspath{{./figures/}}
\newcommand{\architecture}[1]{\emph{#1}}

\begin{document}
\pagestyle{headings}
\mainmatter


\title{Learning to Train a Binary Neural Network} 
\titlerunning{Learning to Train a Binary Neural Network}
\authorrunning{Joseph Bethge, Haojin Yang, Christian Bartz, Christoph Meinel}

\author{Joseph Bethge, Haojin Yang, Christian Bartz, Christoph Meinel}
\institute{
Hasso Plattner Institute, University of Potsdam, Germany \\
P.O. Box 900460, Potsdam D-14480 \\
\{joseph.bethge,haojin.yang,christian.bartz,meinel\}@hpi.de
}

\maketitle


\begin{abstract}
Convolutional neural networks have achieved astonishing results in different application areas.
Various methods which allow us to use these models on mobile and embedded devices have been proposed.
Especially binary neural networks seem to be a promising approach for these devices with low computational power.
However, understanding binary neural networks and training accurate models for practical applications remains a challenge.
In our work, we focus on increasing our understanding of the training process and making it accessible to everyone.
We publish our code and models based on BMXNet for everyone to use\footnote{\url{https://github.com/Jopyth/BMXNet}}.
Within this framework, we systematically evaluated different network architectures and hyperparameters to provide useful insights on how to train a binary neural network.
Further, we present how we improved accuracy by increasing the number of connections in the network.

\end{abstract}


\section{Introduction}
\label{sec:intro}



\begin{table}[t]
\centering
\caption{
Comparison of available implementations for binary neural networks
}
\label{tab:comparison}
\begin{tabular}{|l|c|c|c|c|c|c|c|c|}
\hline
\textbf{Title}                                                                                 & \multicolumn{1}{l|}{\textbf{GPU}} & \multicolumn{1}{l|}{\textbf{CPU}} & \multicolumn{1}{l|}{\textbf{\begin{tabular}[c]{@{}l@{}}Python\\ API\end{tabular}}} & \multicolumn{1}{l|}{\textbf{\begin{tabular}[c]{@{}l@{}}C++\\ API\end{tabular}}} & \multicolumn{1}{l|}{\textbf{\begin{tabular}[c]{@{}l@{}}Save\\ Binary\\ Model\end{tabular}}} & \multicolumn{1}{l|}{\textbf{\begin{tabular}[c]{@{}l@{}}Deploy\\ on\\ Mobile\end{tabular}}} & \multicolumn{1}{l|}{\textbf{\begin{tabular}[c]{@{}l@{}}Open\\ Source\end{tabular}}} & \multicolumn{1}{l|}{\textbf{\begin{tabular}[c]{@{}l@{}}Cross\\ Platform\end{tabular}}} \\ \hline
\begin{tabular}[c]{@{}l@{}}BNNs \cite{Courbariaux2016}  \end{tabular} & \checkmark             &                                   & \checkmark                                                                         &                                                                                 &                                                                                             &                                                                                            & \checkmark                                                                          &                                                                                        \\ \hline
DoReFa-Net \cite{Zhou2016}                                                                     & \checkmark                        & \checkmark                        & \checkmark                                                                         &                                                                                 &                                                                                             &                                                                                            & \checkmark                                                                          & \checkmark                                                                             \\ \hline
XNOR-Net \cite{Rastegari2016}                                                                  & \checkmark                        &                                   &                                                                                    &                                                                                 &                                                                                             &                                                                                            & \checkmark                                                                          &                                                                                        \\ \hline
BMXNet \cite{HPI_xnor}                                                                         & \checkmark                        & \checkmark                        & \checkmark                                                                         & \checkmark                                                                      & \checkmark                                                                                  & \checkmark                                                                                 & \checkmark                                                                          & \checkmark                                                                             \\ \hline
\end{tabular}
\end{table}

Nowadays, significant progress through research is made towards automating different tasks of our everyday lives.
From vacuum robots in our homes to entire production facilities run by robots, many tasks in our world are already highly automated.
Other advances, such as self-driving cars, are currently being developed and depend on strong machine learning solutions.
Further, more and more ordinary devices are equipped with embedded chips (with limited resources) for various reasons, such as smart home devices.
Even operating systems and apps on smartphones adopt deep learning techniques for tackling several problems and will likely continue to do so in the future.
All these devices have limited computational power, often while trying to achieve minimal energy consumption, and might provide future applications for machine learning.

Consider a fully automated voice controlled coffee machine that identifies users by their face and remembers their favorite beverage.
The machine could be connected to a cloud platform which runs the machine learning models and stores user information.
The machine transfers the voice or image data to the server for processing, and receives the action to take or which settings to load.

There are a few requirements for this setup, which can be enumerated easily:
A stable internet connection with sufficient bandwidth is required.
Furthermore, the users have to agree on sending the required data to the company hosting the cloud platform.
This not only requires trust from the users, but data privacy can be an issue, too, especially in other potential application areas, such as healthcare or finances.

All of these potential problems can be avoided by hosting the machine learning models directly on the coffee machine itself.
However, there are other challenges, such as limited computational resources and limited memory, in addition to a possible reliance on battery power.
We focus on solving these challenges by training a Binary Neural Network (BNN).
In a BNN the commonly used full-precision weights of a convolutional neural network are replaced with binary weights.
This results in a storage compression by a factor of 32$\times$ and allows for more efficient inference on CPU-only architectures.
We discuss existing approaches, which have promising results, in \autoref{sec:related}.
However, architectures, design choices, and hyperparameters are often presented without thorough explanation or experiments.
Often, there is no source code for actual BNN implementations present (see \autoref{tab:comparison}).
This makes follow-up experiments and building actual applications based on BNNs difficult.


Therefore we provide our insights on existing network architectures and parameter choices, while striving to achieve a better understanding of BNNs (\autoref{sec:method}).
We evaluate these choices and our novel ideas based on the open source framework BMXNet~\cite{HPI_xnor}.
We discuss the results of a set of experiments on the MNIST, CIFAR10 and ImageNet datasets (\autoref{sec:experiments}).
Finally, we examine future ideas, such as quantized neural networks, wherein the binary weights are replaced with lower precision floating point numbers (\autoref{sec:conclusion}).

Summarized, our contributions presented in this paper are:
\begin{itemize}
    \item We provide novel empirical proof for choice of methods and parameters commonly used to train BNNs, such as how to deal with bottleneck architectures and the gradient clipping threshold. 
    \item We found that dense shortcut connections can improve the classification accuracy of BNNs significantly and show how to create efficient models with this architecture. 
    \item We offer our work as a contribution to the open source framework BMXNet~\cite{HPI_xnor}, from which both academia and industry can take advantage from.
    We share our code and developed models in this paper for research use.
    \item We present an overview about performance of commonly used network architectures with binary weights.
\end{itemize}



\section{Related Work}
\label{sec:related}


In this section we first present two network architectures, Residual Networks \cite{He2017} and Densely Connected Networks \cite{Huang2016}, which focus on increasing information flow through the network.
Afterwards we give an overview about networks and techniques which were designed to allow execution on mobile or embedded devices.

Residual Networks \cite{He2017} combine the information of all previous layers with shortcut connections leading to increased information flow.
This is done through addition of identity connections to the outputs of previous layers together with the output of the current layer.
Consequently, the shortcut connections add neither extra weights nor computational cost.


In Densely Connected Networks \cite{Huang2016} the shortcut connections are instead built by concatenating the outputs of previous layers and the current layer.
Therefore, new information gained in one layer can be reused throughout the entire depth of the network.
To reduce the total model size, the original full-precision architecture includes a bottleneck design, which reduces the number of filters in transition layers.
These effectively keep the network at a very small total size, even though the concatenation adds new information into the network every few layers.

There are two main approaches which allow for execution on mobile devices:
On the one hand, information in a CNN can be compressed through compact network design.
These designs rely on full-precision floating point numbers, but reduce the total number of parameters with a clever network design, while preventing loss of accuracy.
On the other hand, information can be compressed by avoiding the common usage of full-precision floating point weights, which use 32 bit of storage.
Instead, quantized floating-point number with lower precision (e.g. 8 bit of storage) or even binary (1 bit of storage) weights are used in these approaches.


We first present a selection of techniques which utilize the former method.
The first of these approaches, \emph{SqueezeNet}, was presented by Iandola et al.~\cite{Iandola2016} in 2016.
The authors replace a large portion of 3$\times$3 filters with smaller 1$\times$1 filters in convolutional layers and reduce the number of input channels to the remaining 3$\times$3 filters for a reduced number of parameters.
Additionally, they facilitate late downsampling to maximize their accuracy based on the lower number of weights.
Further compression is done by applying \emph{deep compression} \cite{Han2015} to the model for an overall model size of 0.5 MB.

A different approach, \emph{MobileNets}, was implemented by Howard et al.~\cite{Howard2017}.
They use a depth-wise separable convolution where convolutions apply a single 3$\times$3 filter to each input channel.
Then, a 1$\times$1 convolution is applied to combine their outputs.
Zhang et al.~\cite{Zhang2017} use channel shuffling to achieve group convolutions in addition to depth-wise convolution.
Their \emph{ShuffleNet} achieves comparably lower error rate for the same number of operations needed for \emph{MobileNets}.
These approaches reduce memory requirements, but still require GPU hardware for efficient training and inference.
Specific acceleration strategies for CPUs still need to be developed for these methods.


In contrast to this, approaches which use binary weights instead of full-precision weights can achieve compression and acceleration.
However, the drawback usually is a severe drop in accuracy.
For example, the weights and activations in \emph{Binarized Neural Networks} are restricted to either +1 or -1, as presented by Hubara et al.~\cite{Courbariaux2016}.
They further provide efficient calculation methods of the equivalent of a matrix multiplication by using \emph{XNOR} and \emph{popcount} operations.
\emph{XNOR-Nets} are built on a similar idea and were published by Rastegari et al.~\cite{Rastegari2016}.
They include a channel-wise scaling factor to improve approximation of full-precision weights, but require weights between layers to be stored as full-precision numbers.
Another approach, called \emph{DoReFa-Net}, was presented by Zhou et al.~\cite{Zhou2016}.
They focus on quantizing the gradients together with different bit-widths (down to binary values) for weights and activations and replace the channel-wise scaling factor with one constant scalar for all filters.
Another attempt to remove everything except binary weights is taken in \emph{ABC-Nets} by Lin et al.~\cite{lin2017towards}.
This approach achieves a drop in top1-accuracy of only about 5\% on the ImageNet dataset compared to a full-precision network using the \architecture{ResNet} architecture.
They suggest to use between 3 to 5 binary weight bases to approximate full-precision weights, which increases model capacity, but also model complexity and size.
Therefore finding a way to accurately train a binary neural network still remains an unsolved task.



\section{Methodology}
\label{sec:method}


In alignment with our goal to contribute to open-source frameworks, we publish the code and models and offer them as a contribution to the BMXNet framework.
A few implementation details are provided here.
We use the sign function for activation (and thus transform from real-valued values into binary values):
\begin{equation}
    \mathrm{sign}(x) = \begin{cases} 
    +1 ~\text{if}~ x \geq 0, \\
    -1 ~\text{otherwise},
    \end{cases}
\end{equation}
The implementation uses a Straight-Through Estimator (STE) \cite{hinton2012neural} which cancels the gradients, when they get too large, as proposed by Hubara et al.~\cite{Courbariaux2016}.
Let $c$ denote the objective function, $r_i$ be a real number input, and $r_o\in\{-1,+1\}$ a binarized output.
Furthermore $t_{clip}$ is a threshold for clipping gradients.
In previous works the clipping threshold was set to $t_{clip}=1$ \cite{Courbariaux2016}.
Then, the straight-through estimator is:

\begin{align}
    \text{Forward:} ~r_o=\mathrm{sign}(r_i) ~~~~~~ \text{Backward:} ~\frac{\partial c}{\partial r_i}=\frac{\partial c}{\partial r_o}1_{|r_i|\leq t_{clip}}
\end{align}

Usually in full-precision networks a large amount of calculations is spent on calculating dot products of matrices, as is needed by fully connected and convolutional layers.
The computational cost of binary neural networks can be highly reduced by using the \texttt{XNOR} and \texttt{popcount} CPU instructions.
Both operations combined approximate the calculation of dot products of matrices.
That is because element-wise multiplication and addition of a dot product can be replaced with the \texttt{XNOR} instruction and then counting all bits, which are set to 1 (\texttt{popcount}) \cite{Rastegari2016}.
Let $x,w \in \{-1,+1\}^n$ denote the input and weights respectively (with $n$ being the number of inputs).
Then the matrix multiplication $x\cdot w$ can be replaced as follows:
\begin{equation}
\label{eqn:xnor-popcount}
    x \cdot w = 2 \cdot \mathrm{bitcount}(\mathrm{xnor}(x,w)) - n
\end{equation}
Preliminary experiments showed, that an implementation as custom CUDA kernels was slower than using the highly optimized cuDNN implementation.
But the above simplification means, that we can still use normal training methods with GPU acceleration.
We simply need to convert weights from $\{-1,+1\}$ to $\{0,1\}$ before deployment in a CPU architecture.
Afterwards we can take advantage of the CPU implementation.

In the following sections we describe which parameters we evaluate and how we gain explanations about the whole system.
First, we discuss common training parameters, such as including a scaling factor during training and the threshold for clipping the gradients.
Secondly, we examine different deep neural network architectures, such as \architecture{AlexNet} \cite{krizhevsky2012imagenet}, \architecture{Inception} \cite{szegedy2015going,Szegedy2015}, \architecture{ResNet} \cite{He2017}, \architecture{DenseNet} \cite{Huang2016}.
During this examination, we focus on the effect of reducing weights in favor of increasing the number of connections on the example of the \architecture{DenseNet} architecture.
Thirdly, we determine the differences of learned features between binary neural networks and full-precision networks with feature visualization.

\subsection{Network Architectures}

\begin{figure*}[t]
\captionsetup[subfigure]{justification=centering}
\begin{center}
\begin{subfigure}[t]{0.24\linewidth}
   \centering
   \includegraphics[width=0.915\linewidth]{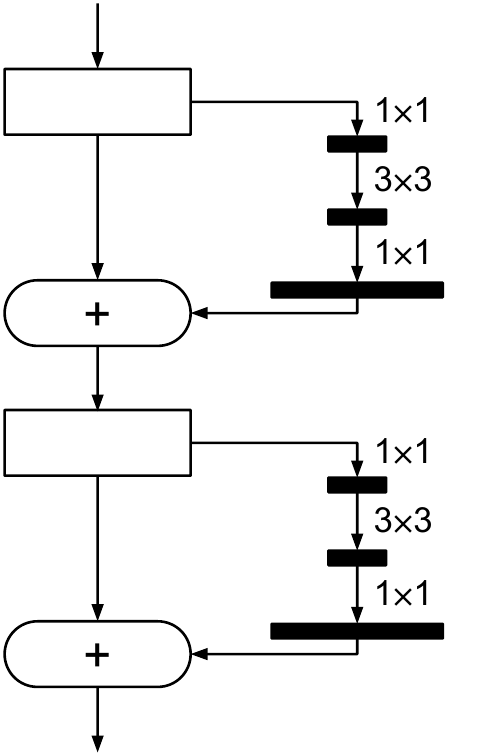}
   \caption{ResNet \\ (bottleneck)}
   \label{fig:netblocks-resnet-bottleneck}
\end{subfigure}
 \begin{subfigure}[t]{0.24\linewidth}
   \centering
   \includegraphics[width=0.915\linewidth]{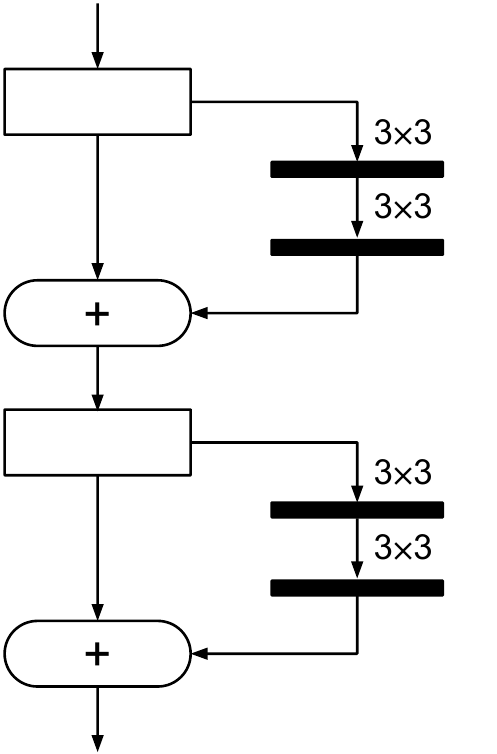}
   \caption{ResNet}
   \label{fig:netblocks-resnet}
\end{subfigure}
\begin{subfigure}[t]{0.24\linewidth}
   \centering
   \includegraphics[width=0.99\linewidth]{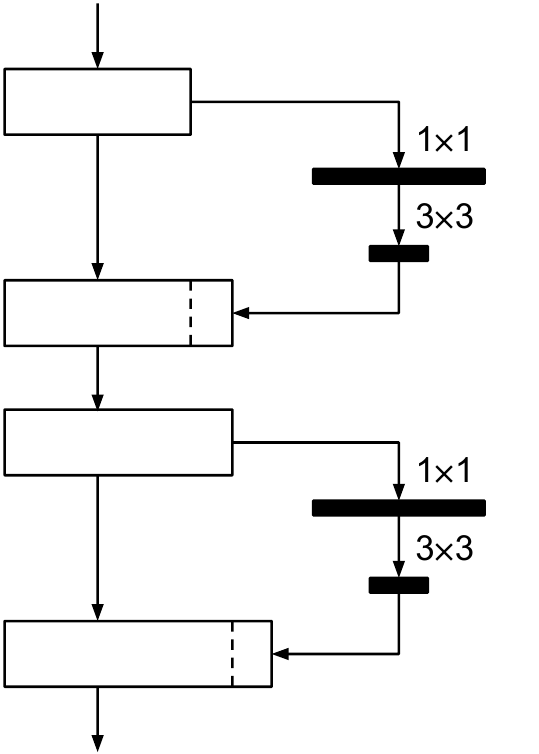}
   \caption{DenseNet \\ (bottleneck)}
   \label{fig:netblocks-densenet-bottleneck}
\end{subfigure}
 \begin{subfigure}[t]{0.24\linewidth}
   \centering
   \includegraphics[width=0.99\linewidth]{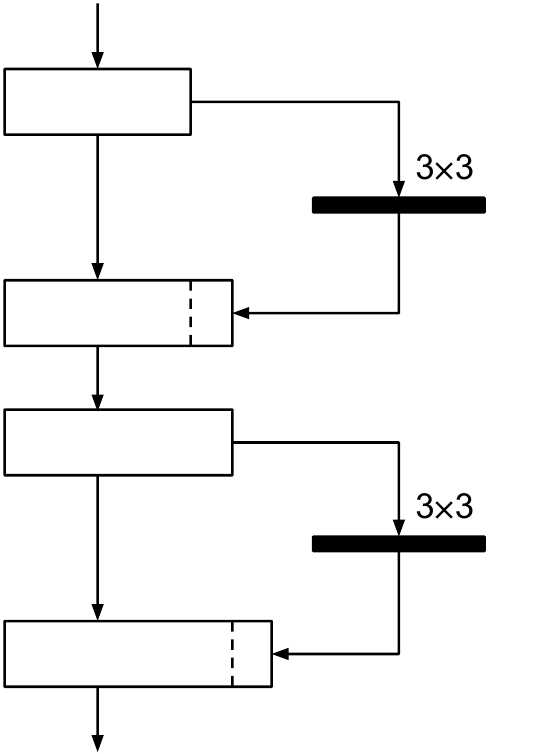}
   \caption{DenseNet}
   \label{fig:netblocks-densenet}
\end{subfigure}
\end{center}
\caption{
    Two (identical) building blocks of different network architectures.
    (\subref{fig:netblocks-resnet-bottleneck}) The original \architecture{ResNet} design features a bottleneck architecture (length of bold black line represents number of filters).
    A low number of filters reduces information capacity for binary neural networks.
    (\subref{fig:netblocks-resnet}) A variation of the \architecture{ResNet} architecture without the bottleneck design.
    The number of filters is increased, but with only two convolutions instead of three.
    (\subref{fig:netblocks-densenet-bottleneck}) The original \architecture{DenseNet} design with a bottleneck in the second convolution operation.
    (\subref{fig:netblocks-densenet}) The \architecture{DenseNet} design without a bottleneck.
    The two convolution operations are replaced by one $3\times3$ convolution
}
\label{fig:netblocks}
\end{figure*}
%


Before thinking about model architectures, we must consider the main aspects, which are necessary for binary neural networks.
First of all, the information density is theoretically 32 times lower, compared to full-precision networks.
Research suggests, that the difference between 32 bits and 8 bits seems to be minimal and 8-bit networks can achieve almost identical accuracy as full-precision networks \cite{Han2015}.
However, when decreasing bit-width to four or even one bit (binary), the accuracy drops significantly \cite{Courbariaux2016}.
Therefore, the precision loss needs to be alleviated through other techniques, for example by increasing information flow through the network.
This can be successfully done through shortcut connections, which allow layers later in the network to access information gained in earlier layers despite of information loss through binarization.
These shortcut connections, were proposed for full-precision model architectures in Residual Networks \cite{He2017} and Densely Connected Networks \cite{Huang2016} (see \autoref{fig:netblocks}\subref{fig:netblocks-resnet-bottleneck}, \subref{fig:netblocks-densenet-bottleneck}).

Following the same idea, network architectures including bottlenecks always are a challenge to adopt.
The bottleneck architecture reduces the number of filters and values significantly between the layers, resulting in less information flow through binary neural networks.
Therefore we hypothesize, that either we need to eliminate the bottleneck parts or at least increase the number of filters in these bottleneck parts for accurate binary neural networks to achieve best results (see \autoref{fig:netblocks}\subref{fig:netblocks-resnet}, \subref{fig:netblocks-densenet}).

\begin{figure*}[t]
\captionsetup[subfigure]{justification=centering}
\begin{center}
\begin{subfigure}[t]{0.32\linewidth}
   \centering
   \includegraphics[width=0.95\linewidth]{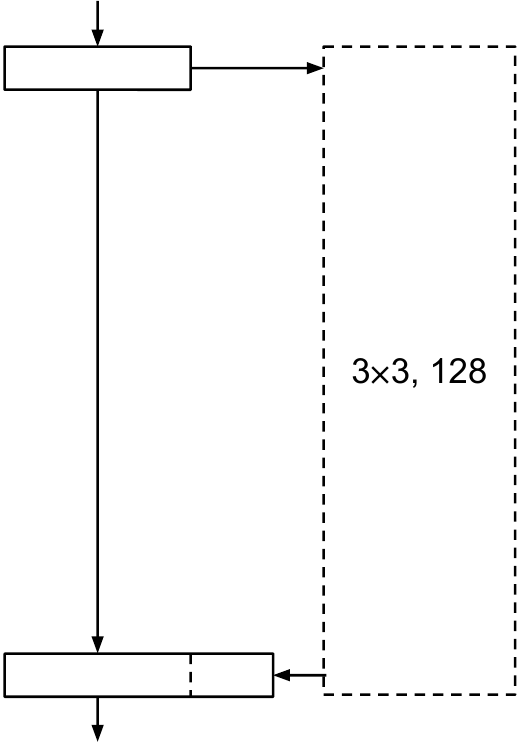}
   \caption{}
   \label{fig:connections-128}
\end{subfigure}
\hfill
\begin{subfigure}[t]{0.32\linewidth}
   \centering
   \includegraphics[width=0.95\linewidth]{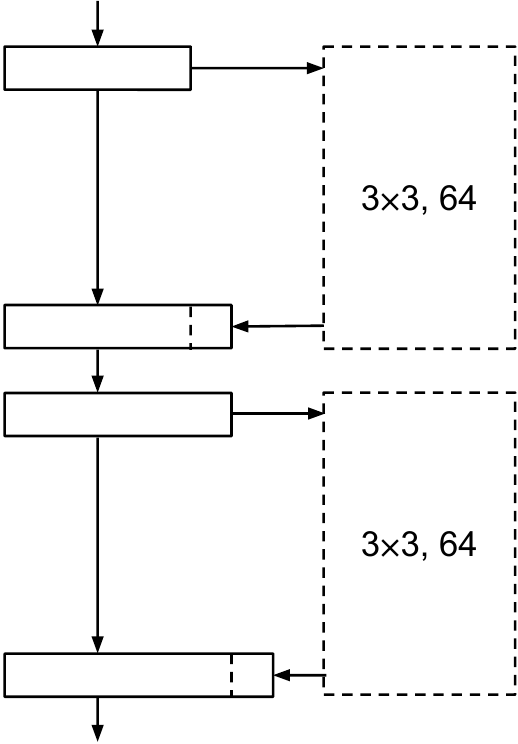}
   \caption{}
   \label{fig:connections-64}
\end{subfigure}
\hfill
\begin{subfigure}[t]{0.32\linewidth}
   \centering
   \includegraphics[width=0.95\linewidth]{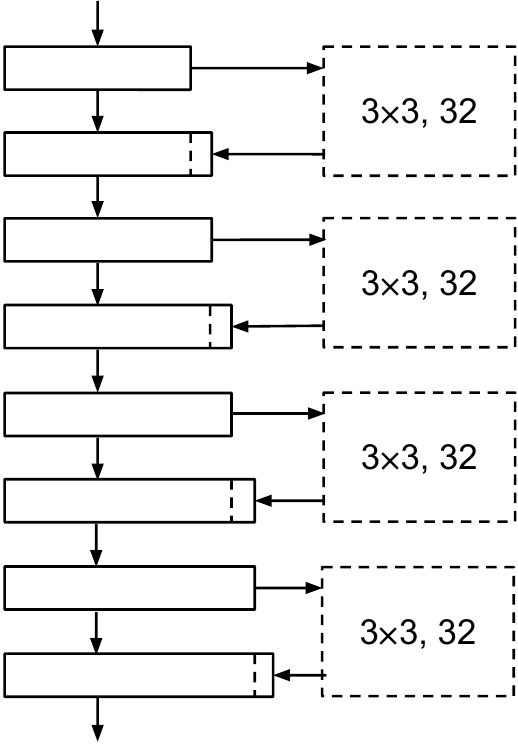}
   \caption{}
   \label{fig:connections-32}
\end{subfigure}
\end{center}
\caption{
  Different ways to extract information with 3$\times$3 convolutions.
  (\subref{fig:connections-128}) A large block which generates a high amount of features through one convolution.
  (\subref{fig:connections-64}) Splitting one large block in two, which are half as large and generate half as many features respectively.
  This allows the features generated in the first block to be used by the second block.
  (\subref{fig:connections-32}) This process can be repeated until a minimal desirable block size is found (e.g. 32 for binary neural networks)
}
\label{fig:connections}
\end{figure*}


To increase the information flow, the blocks which add or derive new features to \architecture{ResNet} and \architecture{DenseNet} (see \autoref{fig:netblocks}) have to be modified.
In full-precision networks, the size of such a block ranges from 64 to 512 for \architecture{ResNet} \cite{He2017}.
The authors of \architecture{DenseNet} call this parameter growth rate and set it to $k=32$ \cite{Huang2016}.
Our preliminary experiments showed, that reusing the full-precision \architecture{DenseNet} architecture for binary neural networks and only removing the bottleneck architecture, is not achieving satisfactory performance.
There are different possibilities to increase the information flow for a \architecture{DenseNet} architecture.
The growth rate can be increased (e.g. $k=64, k=128$), we can use a larger number of blocks, or a combination of both (see \autoref{fig:connections}).
Both approaches add roughly the same amount of parameters to the network.
It is not exactly the same, since other layers also depend on the growth rate parameter (e.g. the first fully-connected layer which also changes the size of the final fully-connected layer and the transition layers).
Our hypothesis of favoring an increased number of connections over simply adding more weights indicates, that in this case increasing the number of blocks should provide better results (or a reduction of the total number of parameters for equal model performance) compared to increasing the growth rate.

\subsection{Common Hyperparameters}
\label{sec:hyperp}


One technique which was used in binary neural networks before, is a scaling factor \cite{Zhou2016,Rastegari2016}.
The result of a convolution operation is multiplied by this scaling factor.
This should help binary weights to act more similarly to full-precision weights, by increasing the value range of the convolution operation.
However, this factor was applied in different ways.
We evaluated whether this scaling factor proves useful in all cases, because it adds additional complexity to the computation and the implementation in \autoref{sec:experiments-acc}.


Another parameter specific to binary neural networks, is the clipping threshold $t_{clip}$.
The value of this parameter influences which gradients are canceled and which are not.
Therefore the parameter has a significant influence on the training result, and we evaluated different values for this parameter (also in \autoref{sec:experiments-acc}).

\subsection{Visualization of Trained Models}


We used an implementation of the \emph{deep dream} visualization \cite{mordvintsev2015} to visualize what the trained models had learned (see \autoref{fig:deepdream}).
The core idea is a normal forward pass followed by specifying an optimization objective, such as maximizing a certain neuron, filter, layer, or class during the backward pass.

Another tool we used for visualization is \emph{VisualBackProp}~\cite{Bojarski2016}.
It uses the high certainty about relevancy of information of the later layers in the network together with the higher resolution of earlier layers to efficiently identify those parts in the image which contribute most to the prediction.



\section{Experiments and Results}
\label{sec:experiments}


Following the structure of the previous section, we provide our experimental results to compare the various parameters and techniques.
First, we focus on classification accuracy as a measure to determine which parameter choices are better.
Afterwards, we examine the results of our feature visualization techniques.

\subsection{Classification Accuracy}
\label{sec:experiments-acc}




In this section we apply classification accuracy as the general measurement to evaluate the different architectures, hyperparameters etc.
We use the MNIST~\cite{lecun-mnisthandwrittendigit-2010}, CIFAR-10~\cite{cifar10} and ImageNet~\cite{imagenet_cvpr09} datasets in terms of different levels of task complexity.
The experiments were performed on a work station which has an Intel(R) Core(TM) i9-7900X CPU, 64 GB RAM and 4$\times$Geforce GTX1080Ti GPUs.

As a general experiment setup, we use full-precision weights for the first (often a convolutional) layer and the last layer (often a fully connected layer which has a number of output neurons equal to the number of classes) for all involved deep networks.
We did not apply a scaling factor as proposed by Rastegari et al. in~\cite{Rastegari2016} in our experiments.
Instead we examined a (similar) scaling factor method proposed by Zhou et al.~\cite{Zhou2016}.
However, as shown in our hyperparameter evaluation (page \pageref{sec:scale-factor}) we chose not to apply this scaling factor for our other experiments.
Further, the results of a binary \emph{LeNet} for the MNIST dataset and a binary \emph{DenseNet} with 21 layers can be seen in \autoref{tab_mnist_cifar}.

\begin{table*}[b]
\caption{Evaluation of model performance on the MNIST and CIFAR-10 data sets.}
\begin{center}
\begin{tabular}{| p{2.6cm} | p{2.4cm} | p{1.5cm} | p{5.0cm} |}
    \hline
                     & Architecture & Accuracy & Model Size (Binary/Full Precision)\\
    \hline
MNIST & LeNet  & \textbf{99.3\%} & 202KB/4.4MB \\
    \hline
CIFAR-10 & DenseNet-21  & \textbf{87.1\%} & 1.9MB/51MB\\ \cline{2-3}    
    \hline 
\end{tabular} 
\end{center}
\label{tab_mnist_cifar}
\end{table*}

\begin{table*}[t]
\caption{
Classification accuracy (Top-1 and Top-5) of several popular deep learning architectures using binary weights and activations in their convolution and fully connected layers.
Full-precision results are denoted with FP.
\architecture{ResNet-34-thin} applies a lower number of filters ($64, 64, 128, 256, 512$), whereas \architecture{ResNet-34-wide} and \architecture{ResNet-68-wide} use a higher number of filters ($64, 128, 256, 512, 1024$).}
\begin{center}
\begin{tabular}{ |c|c|c|c|c|c|c| }
\hline
Architecture & Top-1 & Top-5 & Epoch & \multicolumn{1}{l|}{\begin{tabular}[c]{@{}c@{}}Model Size\\(Binary/Full Precision)\end{tabular}} & \multicolumn{1}{l|}{\begin{tabular}[c]{@{}c@{}}Top-1\\FP\end{tabular}} & \multicolumn{1}{l|}{\begin{tabular}[c]{@{}c@{}}Top-5\\FP\end{tabular}} \\
\hline
AlexNet & 30.2\% &  54.0\% & 70 & 22MB/233MB & 62.5\% & 83.0\% \\
\hline
InceptionBN & 24.8\% & 48.0\% & 80 & 8MB/44MB & - & 92.1\% \\
\hline
ResNet-18 & 42.0\% & 66.2\% & 37 & 3.4MB/45MB & - & - \\
\hline
ResNet-18 (from \cite{lin2017towards}) & 42.7\% & 67.6\% & - & - & - & - \\
\hline
ResNet-26 bottleneck & 25.2\% & 47.1\% & 40 & - & - & - \\
\hline
ResNet-34-thin & 44.3\% & 69.1\% & 40 & 4.8MB/84MB & 78.2\% & 94.3\% \\
\hline
ResNet-34-wide & 54.0\% & 77.2\% & 37 & 15MB/329MB & - & - \\
\hline
ResNet-68-wide & 57.5\% & 80.3\% & 40 & 25MB/635MB & - & - \\
\hline
\end{tabular}
\end{center}
\label{tab_dn_eval1}
\end{table*}

\subsubsection{Popular Deep Architectures}
In this experiment our intention is to evaluate a selection of popular deep learning architectures by using binary weights and activations.
We wanted to discover positive and negative design patterns with respect to training binary neural networks.
The first experiment is based on AlexNet~\cite{krizhevsky2012imagenet}, InceptionBN~\cite{ioffe2015batch} and ResNet~\cite{He2017} (see \autoref{tab_dn_eval1}).
Using the AlexNet architecture, we were not able to achieve similar results as presented by Rastegari et al.~\cite{Rastegari2016}.
This might be due to us disregarding their scaling factor approach.
Further, we were quite surprised that InceptionBN achieved even worse results than AlexNet.
Our assumption for the bad result is that the Inception series applies ``bottleneck'' blocks intended to reduce the number of parameters and computational costs, which may negatively impact information flow.
With this idea, we continued the experiments with several ResNet models, and the results seem to verify our conjecture.
If the \architecture{ResNet} architecture is used for full-precision networks, gradually increasing the width and depth of the network yields improvements in accuracy.
On the contrary, when using binary neural networks, the bottleneck design seems to limit the performance as is expected.
We were not able to obtain higher accuracy with the \architecture{ResNet-26 bottleneck} architecture compared to \architecture{ResNet-18}.
Additionally, if we only increase the depth, without increasing the number of filters, we were not able to obtain a significant increase in accuracy (\architecture{ResNet-34-thin} compared to \architecture{ResNet-18}).
To test our theory, that the bottleneck design hinders information flow, we enlarged the number of filters throughout the network from $(64, 64, 128, 256, 512)$ to $(64, 128, 256, 512, 1024)$.
This achieves almost \textbf{10\%} top-1 accuracy gain in a ResNet architecture with 34 layers (\architecture{ResNet-34-wide}).
Further improvements can be obtained by using \architecture{ResNet-68-wide} with both increased depth and width.
This suggests, that network width and depth should be increased simultaneously for best results.

We also conducted experiments on further architectures such as \architecture{VGG-Net}~\cite{simonyan2014very}, \architecture{Inception-resnet}~\cite{szegedy2017inception} and \architecture{MobileNet}~\cite{Howard2017}.
Although we applied batch normalization, the \architecture{VGG}-style networks with more than 10 layers have to be trained accumulatively (layer by layer),
since the models did not achieve any result when we trained them from scratch.
Other networks such as \architecture{Inception-ResNet} and \architecture{MobileNet} are also not appropriate for the binary training due to their designed architecture (bottleneck design and models with a low number of filters).
We assume that the shortcut connections of the \architecture{ResNet} architecture can retain the information flow unobstructed during the training.
This is why we could directly train a binary \architecture{ResNet} model from scratch without additional support.
According to the confidence of the results obtained in our experiment, we achieved the same level in terms of classification accuracy comparing to the latest result from \architecture{ABC-Net}~\cite{lin2017towards} (\architecture{ResNet-18} result with weight base 1 and activation base 1).

As we learned from the previous experiments, we consider the shortcut connections as a useful compensation for the reduced information flow.
But this raised the following question: could we improve the model performance further by simply increasing the number of shortcut connections?
To answer this question, we conducted further experiments based on the \architecture{DenseNet}~\cite{Huang2016} architecture.

%
\begin{table*}[t]
\caption{
Classification accuracy comparison by using binary DenseNet and ResNet models for the ImageNet dataset. 
The amount of parameters are kept on a similar level for both architectures to verify that improvements are based on the increased number of connections and not an increase of parameters.
}
\begin{center}
\begin{tabular}{ |c|c|c|c|c|c| }
\hline
Architecture & Top-1 & Top-5 & Epoch & Model Size (FP) & Number of Parameters\\
\hline
DenseNet-21 & 50.0\% & 73.3\% & 49 & 44MB & 11 498 086\\
\hline
ResNet-18 & 42.0\% & 66.2\% & 37 & 45MB & 11 691 950\\
\hline
DenseNet-45 & 57.9\% & 80.0\% & 52 & 250MB & 62 611 886\\
\hline
ResNet-34-wide & 54.0\% & 77.2\% & 37 & 329MB & 86 049 198\\
\hline
ResNet-68-wide & 57.5\% & 80.3\% & 35 & 635MB & 166 283 182\\
\hline
\end{tabular}
\end{center}
\label{tab_dense_res_comp}
\end{table*}

\subsubsection{Shortcut Connections Driven Accuracy Gain}
In our first experiment we created binary models using both \architecture{DenseNet} and \architecture{ResNet} architectures with similar complexities.
We keep the amount of parameters on a roughly equal level to verify that the improvements obtained by using the \architecture{DenseNet} architecture are coming from the increased number of connections and not a general increase of parameters.
Our evaluation results show that these dense connections can significantly compensate for the information loss from binarization (see \autoref{tab_dense_res_comp}).
The gained improvement by using \architecture{DenseNet-21} compared to \architecture{ResNet-18} is up to \textbf{8\%}\footnote{We note, that this is significantly more, than the improvement between two full-precision models with a similar number of parameters (\emph{DenseNet-264} and \emph{ResNet-50}), which is less than \textbf{2\%} (22.15\% and 23.9\% top 1 error rate, reported by \cite{Huang2016}).}, whereas the number of utilized parameters is even lower.
Furthermore, when we compare binary \architecture{ResNet-68-wide} to \architecture{DenseNet-45}, the latter has less than half the number of parameters compared to the former, but can achieve a very similar result in terms of classification accuracy.

\begin{figure*}[t]
\includegraphics[width=\linewidth]{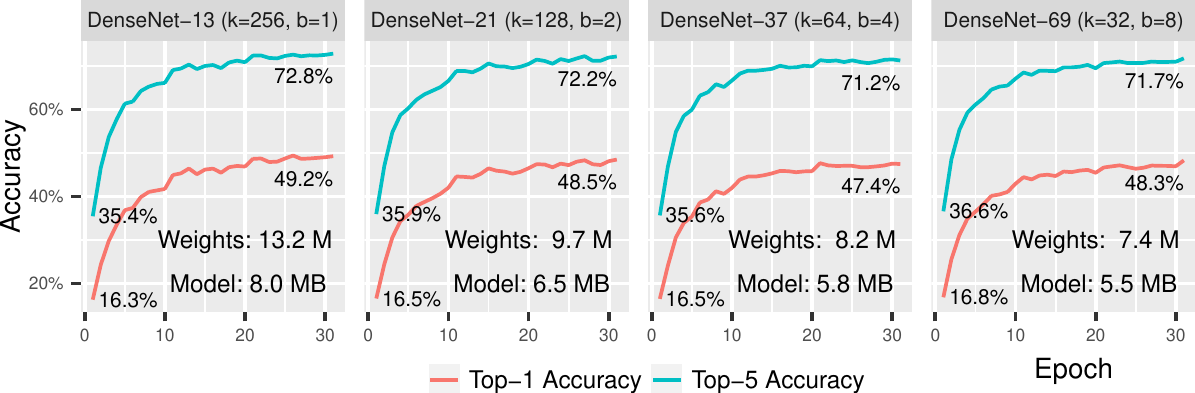}
\caption{
Model performance of binary \architecture{DenseNet} models with different growth rates $k$ and number of blocks $b$.
Increasing $b$, while decreasing $k$ leads to smaller models, without a significant decrease in accuracy, since the reduction of weights is compensated by increasing the number of connections
}
\label{fig:connection-results}
\end{figure*}

In our second set of experiments, we wanted to confirm our hypothesis, that increasing the number of blocks is more efficient than just increasing block size on the example of a \architecture{DenseNet} architecture.
We distinguish the four architectures through the two main parameters relevant for this experiment: growth rate $k$ per block, number of blocks per unit $b$, and total number of layers $n$, where $n = 8 \cdot b + 5$.
The four architectures we are comparing are:
\architecture{DenseNet-13} ($k=256$, $b=1$),
\architecture{DenseNet-21} ($k=128$, $b=2$),
\architecture{DenseNet-37} ($k=64$, $b=4$), and
\architecture{DenseNet-69} ($k=32$, $b=8$).

Despite a 31\% reduction in model size between \architecture{DenseNet-13} and \architecture{DenseNet-69}, the accuracy loss is only 1\% (see \autoref{fig:connection-results}).
We further conclude, that this similarity is not randomness, since all architectures perform very similarly over the whole training process.
We note again, that for all models the first convolutional layer and the final fully-connected layer use full-precision weights.
Further, we set the size of the former layer depending on the growth rate $k$, with a number of filters equal to $2\cdot k$.
Therefore, a large portion of the model size reduction comes from reducing the size of first convolutional layer, which subsequently also reduces the size of the final fully connected layer.

However, a larger fully-connected layer could simply add additional duplicate or similar features, without affecting performance.
This would mean, that the reduction of model size in our experiments comes from a different independent variable.
To elimnate this possibility, we ran a post-hoc analysis to check whether we can reduce the size of the first layer without impacting performance.
We used \architecture{DenseNet-13} with a reduced first layer, which has the same size as for \architecture{DenseNet-69} (which uses $k=32$), so $2 \cdot k = 64$ filters.
Even though the performance of the model is similar for the first few epochs, the accuracy does not reach comparable levels:
after 31 Epochs, its Top-1 accuracy is only 47.1\% (2.1\% lower) and its Top-5 accuracy is only 70.7\% (2.1\% lower).
In addition to degrading the accuracy more than as if increasing connections, it only reduces the model size by 6\% (0.4 MB), since the transition layers are unchanged.
This confirms our hypothesis, that we can eliminate the usual reduction in accuracy of a binary neural network when reducing the number of weights by increasing the number of connections.

In summary, we have learned two important findings from the previous experiments for training an accurate binary network: 
\begin{itemize}
  \item Increasing information flow through the network improves classification accuracy of a binary neural network.
  \item We found two ways to realize this: Increase the network width appropriately while increasing depth or increasing the number of shortcut connections.
\end{itemize}

\subsubsection{Specific Hyperparameter Evaluation}

In this section we evaluated two specific hyperparameters for training a binary neural network: the gradient clipping threshold and usage of a scaling factor.
\begin{figure*}[!t]
\begin{center}
\begin{subfigure}[t]{0.475\linewidth}
   \centering
   \includegraphics[width=\linewidth]{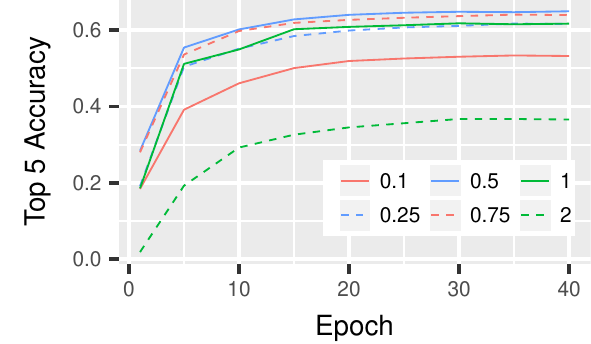}
   \caption{}
   \label{fig_gradient_clipping_threshold}
\end{subfigure}
\begin{subfigure}[t]{0.475\linewidth}
   \centering
   \includegraphics[width=\linewidth]{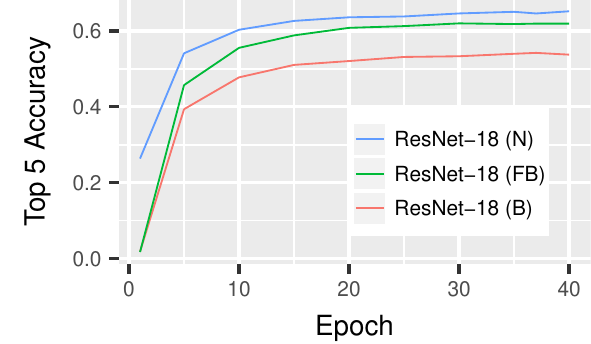}
   \caption{}
   \label{fig_scaling_scalar}
\end{subfigure}
\end{center}
\caption[Efficiency analysis of xnor-dot engine]{
(\subref{fig_gradient_clipping_threshold}) Classification accuracy by varying gradient clipping threshold.
The applied validation model is trained on ImageNet with \architecture{ResNet-18} topology.
(\subref{fig_scaling_scalar}) Accuracy evaluation by using scaling factor on network weights in three different modes: (\emph{N}) no scaling, (\emph{B}) use scaling factor on weights only in backward computation, and (\emph{FB}) apply weight scaling in both forward and backward pass.
}
\end{figure*}

Using a \textbf{gradient clipping threshold $t_{clip}$} was originally proposed by Hubara et al.~\cite{Courbariaux2016}, and reused in more recent work~\cite{Rastegari2016,lin2017towards} (see~\autoref{sec:hyperp}).
In short, when using STE we only let the gradients pass through if the input $r_i$ satisfies $|r_i| \leq t_{clip}$.
Setting $t_{clip} = 1$ is presented in the literature with only cursory explanation.
Thus, we evaluated it by exploring a proper value range (see \autoref{fig_gradient_clipping_threshold}).
We used classification accuracy as the evaluation metric and selected thresholds from the value range of $[0.1, 2.0]$ empirically. 
The validation model is trained on the ImageNet dataset with the \architecture{ResNet-18} network architecture.
From the results we can recognize that $t_{clip} = 1$ is suboptimal, the optimum is between 0.5 and 0.75.
We thus applied $t_{clip} = 0.5$ to the all other experiments in this paper.

\textbf{Scaling factors} have been proposed by Rastegari et al.~\cite{Rastegari2016}.
\label{sec:scale-factor}
In their work, the scaling factor is the mean of absolute values of each output channel of weights.
Subsequently, Zhou et al.~\cite{Zhou2016} proposed a scaling factor, which is intended to scale all filters instead of performing channel-wise scaling.
The intuition behind both methods is to increase the value range of weights with the intention of solving the information loss problem during training of a binary network.
We conducted an evaluation of accuracy according to three running modes according to the implementation of Zhou et al.~\cite{Zhou2016}: (N) no scaling, (B) use the scaling factor on weights only in backward computation, (FB) apply weight scaling in both forward and backward pass.
The result indicates that no accuracy gain can be obtained by using a scaling factor on the \architecture{ResNet-18} network architecture (see \autoref{fig_scaling_scalar}).
Therefore we did not apply a scaling factor in our other experiments.

\subsection{Visualization Results}

\begin{figure*}[t]
\begin{center}
\begin{subfigure}[]{0.45\linewidth}
    \begin{minipage}[b]{\linewidth}
    \centering
        \includegraphics[width=0.45\linewidth]{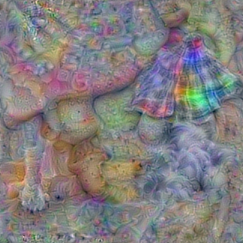}
        \includegraphics[width=0.45\linewidth]{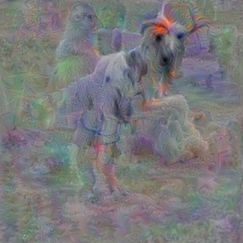}
        \caption{DenseNet (FP)}
        \label{fig:deepdream-densenet-fp}
    \end{minipage}
\end{subfigure}
\begin{subfigure}[]{0.45\linewidth}
    \begin{minipage}[b]{\linewidth}
    \centering
        \includegraphics[width=0.45\linewidth]{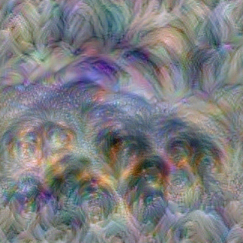}
        \includegraphics[width=0.45\linewidth]{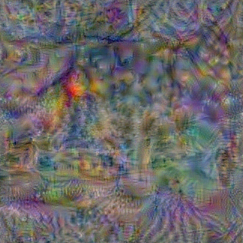}
        \caption{DenseNet-21}
        \label{fig:deepdream-densenet-21}
    \end{minipage}
\end{subfigure}
\begin{subfigure}[]{0.45\linewidth}
    \begin{minipage}[b]{\linewidth}
    \centering
        \includegraphics[width=0.45\linewidth]{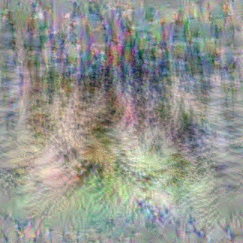}
        \includegraphics[width=0.45\linewidth]{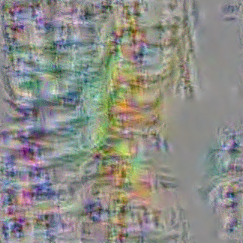}
        \caption{ResNet-18}
        \label{fig:deepdream-resnet-18}
    \end{minipage}
\end{subfigure}
\begin{subfigure}[]{0.45\linewidth}
    \begin{minipage}[b]{\linewidth}
    \centering
        \includegraphics[width=0.45\linewidth]{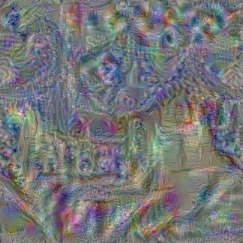}
        \includegraphics[width=0.45\linewidth]{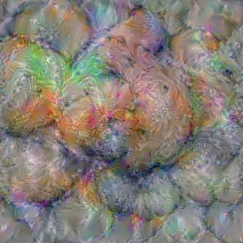}
        \caption{ResNet-68}
        \label{fig:deepdream-resnext-101}
    \end{minipage}
\end{subfigure}
\end{center}
\caption{
  The \emph{deep dream} \cite{mordvintsev2015} visualization of binary models with different complexity and size (best viewed digitally with zoom).
  The \emph{DenseNet} full precision model (\subref{fig:deepdream-densenet-fp}) is the only one, which produces visualizations of animal faces and objects.
  Additional models and samples can be seen in the supplementary material
}
\label{fig:deepdream}
\end{figure*}

\begin{figure*}[t]
\begin{center}
\includegraphics[width=0.49\linewidth]{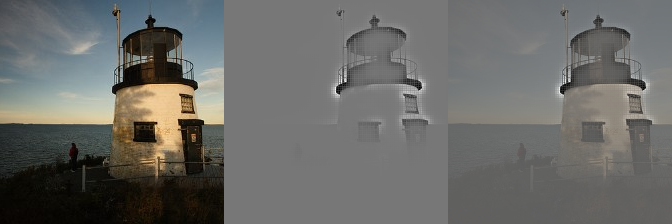}
\includegraphics[width=0.49\linewidth]{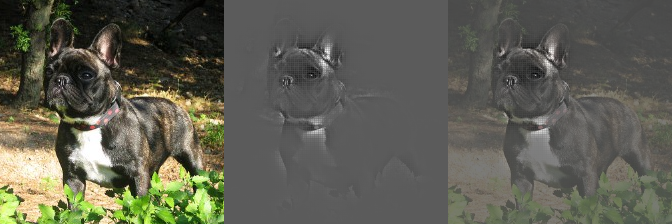}
\includegraphics[width=0.49\linewidth]{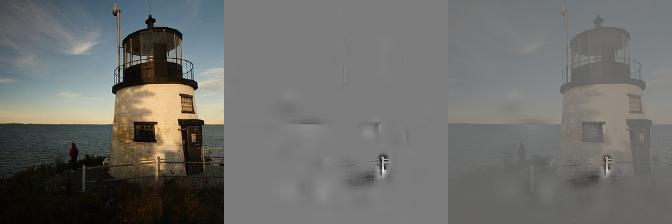}
\includegraphics[width=0.49\linewidth]{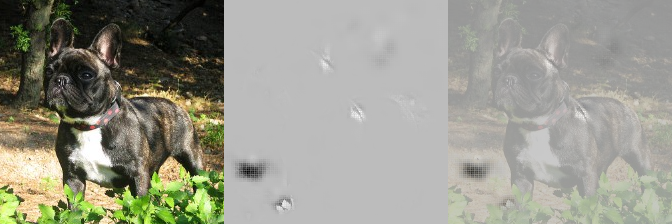}
\includegraphics[width=0.49\linewidth]{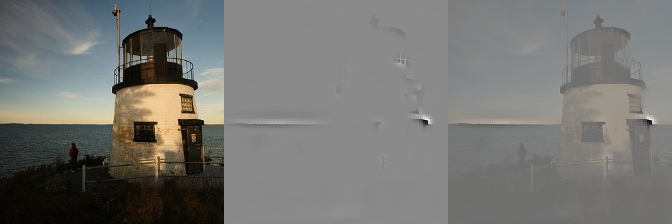}
\includegraphics[width=0.49\linewidth]{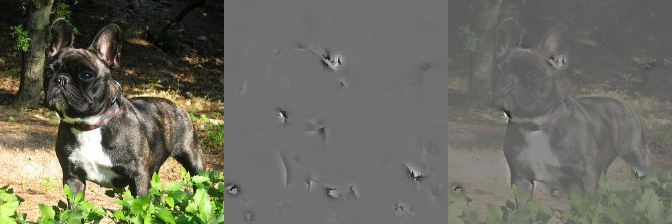}
\end{center}
\caption{
  Two samples of the \emph{ImageNet} dataset visualized with \emph{VisualBackProp} of binary neural network architectures (from top to bottom): full-precision \architecture{ResNet-18}, binary \architecture{ResNet-18}, binary \architecture{DenseNet-21}.
  Each depiction shows (from left to right): original image, activation map, composite of both (best viewed digitally with zoom).
  Additional samples can be seen in the supplementary material
}
\label{fig:visbackprop}
\end{figure*}

To better understand the differences between binary and full-precision networks, and the various binary architectures, we created several visualizations.
The results show, that the full-precision version of the \emph{DenseNet} captures overall concepts, since rough objects, such as animal faces, can be recognized in the \emph{DeepDream} visualization (see \autoref{fig:deepdream-densenet-fp}).
The binary networks perform much worse.
Especially the \emph{ResNet} architecture (see \autoref{fig:deepdream-resnet-18}) with 18 layers seems to learn much more noisy and less coherent shapes.
Further, we can see small and large areas of gray, which hints at the missing information flow in certain parts of the network.
This most likely comes from the loss of information through binarization which stops neurons from activating.
This issue is less visible for a larger architecture, but even there, small areas of gray appear (see \autoref{fig:deepdream-resnext-101}).
However the \emph{DenseNet} architecture (see \autoref{fig:deepdream-densenet-21}) with 21 layers, which has a comparable number of parameters, produces more object-like pictures with less noise.
The areas without any activations seem to not exist, indicating that the information can be passed through the network more efficiently in a binary neural network.

The visualization with \emph{VisualBackprop} shows a similar difference in quality of the learned features (see \autoref{fig:visbackprop}).
It reflects the parts of the image, which contributed to the final prediction of the model.
The visualization of a full-precision \architecture{ResNet-18} clearly highlights the remarkable features of the classes to be detected (e.g. the outline of lighthouse, or the head of a dog).
In contrast, the visualization of a binary \architecture{ResNet-18} only highlights small relevant parts of the image, and considers other less relevant elements in the image (e.g. a horizon behind a lighthouse).
The binary \emph{DenseNet-21} model also achieves less clarity than the full-precision model, but highlights more of the relevant features (e.g. parts of the outline of a dog).


\section{Conclusion}
\label{sec:conclusion}


In this paper, we presented our insights on training binary neural networks.
Our aim is to fill the information gap between theoretically designing binary neural networks, by communicating our insights in this work and providing access to our code and models, which can be used on mobile and embedded devices.
We evaluated hyperparameters, network architectures and different methods of training a binary neural network.
Our results indicate, that increasing the number of connections between layers of a binary neural network can improve its accuracy in a more efficient way than simply adding more weights.


Based on these results, we would like to explore more methods of increasing the number of connections in binary neural networks in our future work.
Additionally similar ideas for quantized networks can be explored, for example, how networks with multiple binary bases work in comparison to quantized low bit-width networks.
The information density should be equal in theory, but are there differences in practice, when training these networks?

\bibliographystyle{splncs}
\bibliography{library}

\end{document}